\documentclass[3p,times]{elsarticle}
\usepackage{csquotes}
\usepackage{multirow}
\usepackage{amssymb}
\usepackage{amsthm}

\usepackage[figuresright]{rotating}

\usepackage{amsmath}
\usepackage[ruled,vlined]{algorithm2e}

\usepackage{hyperref}
\usepackage{xcolor}

\makeatletter
\def\ps@pprintTitle{%
    \let\@oddhead\@empty
    \let\@evenhead\@empty
    \def\@oddfoot{}%
    \let\@evenfoot\@oddfoot}
\makeatother

\begin{document}
\pagestyle{empty}
\begin{frontmatter}




\title{Development of Automated Neural Network Prediction for Echocardiographic Left Ventricular Ejection Fraction}


\author{Yuting Zhang$^{1}$, Boyang Liu$^{2}$, Karina V Bunting$^{3,4}$, David Brind$^{5,6,7}$, Alexander Thorley$^{1}$, \\Andreas Karwath$^{5,7}$,  Wenqi Lu$^{8}$, Diwei Zhou$^{9}$, Xiaoxia Wang$^{3,4,6}$, 
    Alastair R Mobley$^{3,4}$, \\ Otilia Tica$^{3}$, Georgios V Gkoutos$^{5,6,7}$, 
    Dipak Kotecha$^{3,4,6}$, and Jinming Duan$^{1,*}$ 
    \vspace{10pt}
\\ $^{1}$ School of Computer Science, University of Birmingham, Birmingham, UK
\\ $^{2}$ Manchester University NHS Foundation Trust, Manchester, UK
\\ $^{3}$ Institute of Cardiovascular Sciences, University of Birmingham, Edgbaston, Birmingham, UK
\\ $^{4}$ NIHR Birmingham Biomedical Research Centre and West Midlands NHS Secure Data Environment, University Hospitals Birmingham NHS Foundation Trust, Birmingham, UK
\\ $^{5}$ Institute of Cancer and Genomic Sciences, University of Birmingham, Edgbaston, Birmingham, UK
\\ $^{6}$ Health Data Research UK Midlands, University Hospitals Birmingham NHS Foundation Trust, Birmingham, UK
\\ $^{7}$ Centre for Health Data Science, University of Birmingham, Edgbaston, Birmingham, UK
\\ $^{8}$ Department of Computing and Mathematics, Manchester Metropolitan University, Manchester, UK
\\ $^{9}$ Department of Mathematical Sciences, Loughborough University, Loughborough, UK
\\ on behalf of the card\textit{AI}c group}

\address{}

\begin{abstract}
The echocardiographic measurement of left ventricular ejection fraction (LVEF) is fundamental to the diagnosis and classification of patients with heart failure (HF). In order to quantify LVEF automatically and accurately, this paper proposes a new pipeline method based on deep neural networks and ensemble learning. Within the pipeline, an Atrous Convolutional Neural Network (ACNN) was first trained to segment the left ventricle (LV), before employing the area-length formulation based on the ellipsoid single-plane model to calculate LVEF values. This formulation required inputs of LV area, derived from segmentation using an improved Jeffrey’s method, as well as LV length, derived from a novel ensemble learning model. To further improve the pipeline’s accuracy, an automated peak detection algorithm was used to identify end-diastolic and end-systolic frames, avoiding issues with human error. Subsequently, single-beat LVEF values were averaged across all cardiac cycles to obtain the final LVEF. This method was developed and internally validated in an open-source dataset containing 10,030 echocardiograms. The Pearson’s correlation coefficient was 0.83 for LVEF prediction compared to expert human analysis (p$<$0.001), with a subsequent area under the receiver operator curve (AUROC) of 0.98 (95\% confidence interval 0.97 to 0.99) for categorisation of HF with reduced ejection (HFrEF; LVEF$<$40\%). In an external dataset with 200 echocardiograms, this method achieved an AUC of 0.90 (95\% confidence interval 0.88 to 0.91) for HFrEF assessment. This study demonstrates that an automated neural network-based calculation of LVEF is comparable to expert clinicians performing time-consuming, frame-by-frame manual evaluation of cardiac systolic function.  \\
\end{abstract}


\end{frontmatter}


\section{Introduction}
\noindent Heart failure (HF) is a common and increasingly prevalent condition that results in profound burdens on patients, healthcare services and society  \cite{savarese2022global}. It is not a single pathological diagnosis, but rather a clinical syndrome consisting of cardinal symptoms, typical signs on clinical examination, and evidence of impairment of either systolic or diastolic function on cardiac imaging \cite{mcdonagh20212021}. HF is divided into distinct phenotypes based primarily on measurement of systolic left ventricular ejection fraction (LVEF): HF with reduced LVEF (HFrEF, LVEF$<$40\%); HF with mildly reduced ejection fraction (HFmrEF, LVEF 40$-$49\%); and HF with preserved ejection fraction (HFpEF, LVEF $>=$50\%) \cite{mcdonagh20212021,cleland2018beta}. \\

\noindent Echocardiography is one of the most widely utilised diagnostic techniques in cardiology, and is the first-line imaging modality for suspected cardiac pathology due to its availability and portability. The standard method to quantify LVEF using echocardiography as per recommendations from the American Society of Echocardiography (ASE) and the European Association of Cardiovascular Imaging (EACVI), is to first calculate left-ventricular end-diastolic volumes (LVEDV) and end-systolic volumes (LVESV) using the Simpson’s biplane method of multiple discs \cite{lang2015recommendations, lang2005recommendations}. Practically, this method requires sonographers or cardiologists to identify LV end-diastolic (ED) and end-systolic (ES) frames visually from a given cine video, which is both time-consuming and prone to error. There is significant intra- and inter-observer variability in LVEF quantification as a result of poor image quality (the endocardial boarder is often not well seen), and variable cardiac cycle lengths, for example due to arrhythmias such as atrial fibrillation (AF) \cite{myhr2018semi,phad2020left}. To ensure reproducible measurements of LVEF are obtained, it is recommended to average three cardiac cycles for patients in sinus rhythm and five to ten cardiac cycles in AF. These recommendations require substantial training, are rarely followed in clinical practice and are based on consensus opinion only; the available data show that even best practice is time-consuming and poorly reproducible \cite{lang2015recommendations, bunting2021improving}. \\

\noindent To make the calculation of LVEF more efficient and accurate, this paper makes four novel contributions: (1) proposing a new pipeline method to provide comprehensive, transparent details on the calculation of LVEF, which might be more acceptable to clinicians and cardiologists \cite{moal2022explicit}; (2) follow the recommendation by the ASE and EACVI to average LVEF values across all automatically identified cycles for each apical 4-chamber (A4C) echocardiogram; (3) visualise the LV across the full cardiac cycle in a given echocardiogram, thereby useful as an instantaneous summary of beat-to-beat volumetric differences, including the impact of arrhythmias such as AF \cite{sartipy2017atrial, taniguchi2020heart}; and (4) the capacity to predict highly accurate LVEF values at scale without relying on manual approaches that have high workforce requirements.

\section{Methods}
\noindent This project used an overall framework of transparency, as developed by the cardAIc group (Application of Artificial Intelligence to Routine Healthcare Data to Benefit Patients with Cardiovascular Disease) and the BigData@Heart Consortium \cite{savarese2022global}. Reporting follows the DECIDE-AI approach for clinical evaluation of decision support systems driven by artificial intelligence (see supplementary file for DECIDE-AI checklist) \cite{vasey2022reporting,ouyang2019echonet}.

\subsection{Datasets}
\begin{figure}
\centering
\includegraphics[width=0.98\textwidth]{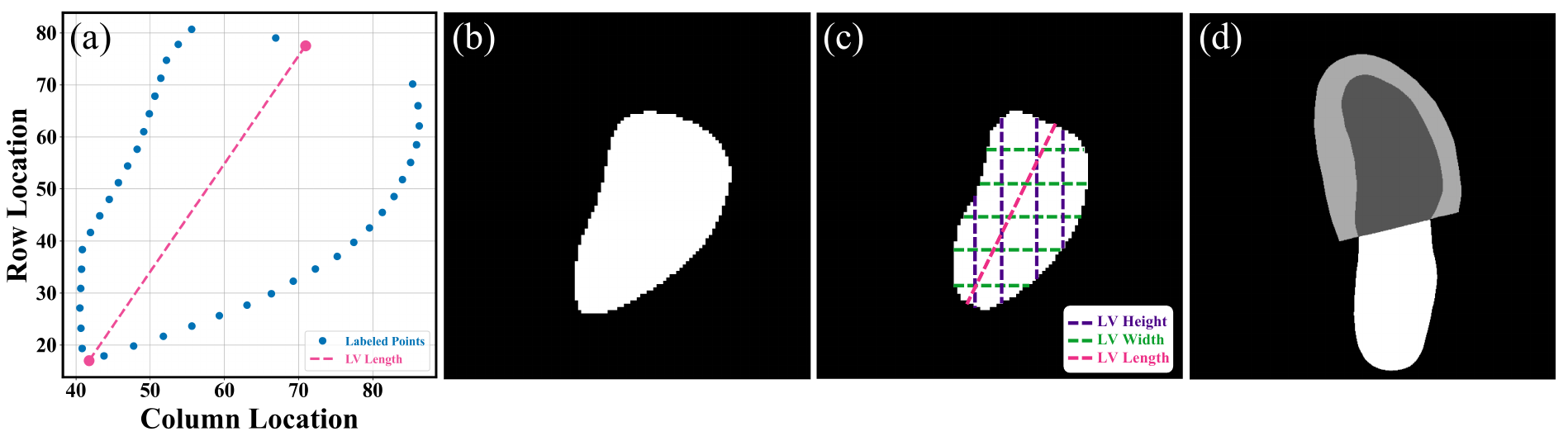}
\caption{(a), (b) and (c) were from the Stanford dataset; (d) was from the CAMUS dataset. (a) presented human labelled coordinate points in one frame. A Euclidean distance between two pink points was the LV length; (b) shown the mask generated from these coordinate points, which was used for training our segmentation network; (c) illustrated the LV area, LV widths, LV heights and LV length; (d) presented annotations information including the left ventricle endocardium, the left ventricle myocardium and the left atrium.}
\label{fig:1}
\end{figure}

\noindent Two open datasets were used in this project, and both of them have obtained ethical approval \cite{ouyang2019echonet,leclerc2019deep}. One is the Stanford dataset with 10,030 A4C 2D gray-scale echocardiogram videos, each of which represented a unique individual who underwent echocardiogram between 2006 and 2018 as part of clinical care; another one is the CAMUS dataset with 450 A4C view sequences, acquired with different ultrasound scanners at the University Hospital of St Etienne (France). For both datasets, labels for each video included the location of the left ventricle endocardium (Fig~\ref{fig:1}(a) and Fig~\ref{fig:1}(d)), LVEF, LVESV, LVEDV, were given by cardiologist experts in the standard clinical workflow. Note that the estimation of left ventricle ejection fraction values was based on the Simpson’s biplane method of discs. For the Stanford dataset, the left ventricle endocardium in ED or ES frames were marked with 42 coordinates, as shown in Fig~\ref{fig:1}(a). More details were supplied in the Appendix B of the supplementary file.

\subsection{AI system }

\begin{figure}
\centering
\includegraphics[width=0.98\textwidth]{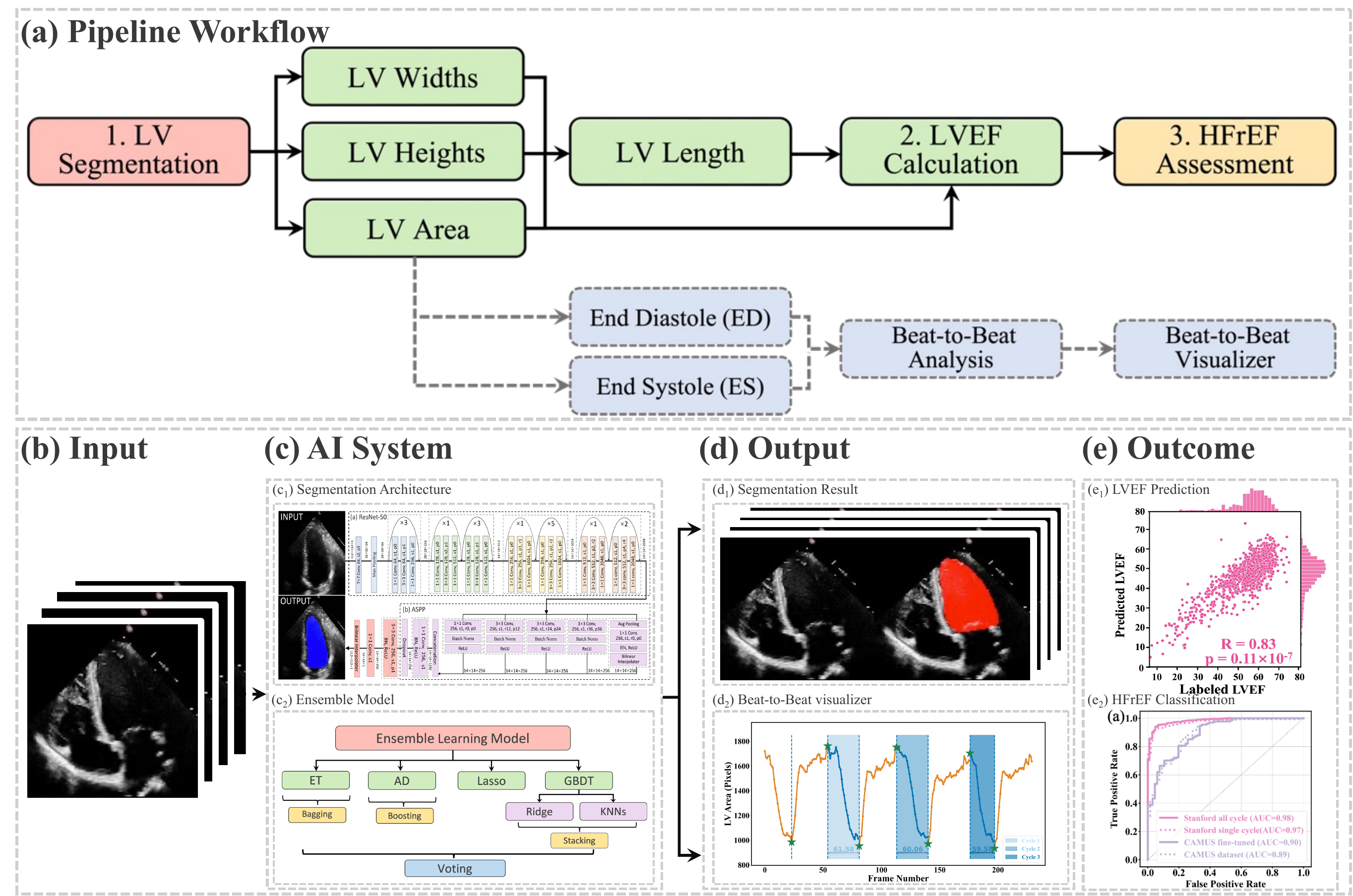}
\caption{(a) Flow chart of the pipeline. There were three main steps, including LV segmentation, LVEF calculation and HFrEF assessment. The area information from segmentation could also be used for ED and ES identification, beat-to-beat analysis of the heart, as well as visualising changes in volume (for example due to an arrhythmia such as atrial fibrillation). ED $=$ end diastole; ES $=$ end systole; HFrEF$=$ Heart failure with reduced LVEF; LV $=$ left ventricle; LVEF$=$ left ventricular ejection fraction. (b) was the input of the pipeline. (c) was the proposed AI system. (d) was outputs information, including the segmentation result and the beat-to-beat visualizer. The calculated LVEF values was presented in this visualizer along with the result of the HF phenotype classification. (e) was the outcome.}
\label{fig:2}
\end{figure}

\noindent \textbf{Methodology:} In this project, the proposed pipeline consisted of three steps to assess patients with HFrEF using their corresponding echocardiogram cine in the A4C view (Fig~\ref{fig:2}(a)). First, an atrous convolutional neural network (ACNN) was employed to segment the LV in each frame of a given video. Based on the segmentation mask, information as shown in Fig~\ref{fig:1}(c), including LV area, LV width, and LV height, was extracted. In addition to segmentation, all ED and ES frames were identified in each video for further beat-to-beat analysis. Second, with the results computed from Step 1, an ensemble learning model was developed to predict the LV length, which was then combined with the LV area to compute LV volumes at ED and ES frames. Based on these LV volumes, the final LVEF was computed (see formulae below). Next, whether a patient has HFrEF was determined based on the LVEF value from Step 2, defined as LVEF $<$40\% \cite{mcdonagh20212021, cleland2018beta}. In addition, a Beat-to-Beat visualizer was provided based on segmentation results to provide an instantaneous summary of beat-to-beat volumetric differences as a result of the heart rhythm. \\

\noindent \textbf{Inputs and Outputs:} The segmentation model required frames or arrays as input, as shown in Fig~\ref{fig:2}(b), with a size of 112x112. Therefore, data preprocessing was carried out before training the pipeline, as described in Appendix B. This pipeline could generate two kinds of outputs, as shown in Fig~\ref{fig:2}(c). One was the segmentation results, which would be displayed in the video format for cardiologists to visualise LV areas across the full cardiac cycle in a given echocardiogram. Another one was the beat-to-beat visualizer, which could be used to visualize how the heart beats and present the LVEF values for each cardiac cycle, along with their average for all cycles. Moreover, based on the LVEF from the all-cycle method, the result of HF phenotype classification was presented in the visualizer. 

\subsection{Implementation}
\noindent In the proposed pipeline, the ellipsoid single-plane model (area-length method) was used to calculate LVEF \cite{stamm1982two}, which was defined in ~\ref{eq:1}.

\begin{equation}
v = \frac{8\pi}{3} \times \frac{A^2}{L}
\label{eq:1}
\end{equation}

In (1), A denoted the LV area, L represented the LV length (the distance from apex to midpoint of the annular plane), and V stood for the volume of LV. With this equation, it was possible to compute the end-diastolic volume (EDV) and end-systolic volume (ESV) of the LV, based on which LVEF is calculated as follows:

\begin{equation}
\text{LVEF} = \frac{1}{N} \sum_{i=1}^{N} \frac{(\text{ESV}_i - \text{EDV}_i)}{\text{EDV}_i} \times 100\%
\label{eq:2}
\end{equation}
Note that information from all cardiac cycles was used and that N here was the available number of cardiac cycles in a video. 
\subsubsection{LV Area}
\begin{figure}
\centering
\includegraphics[width=0.98\textwidth]{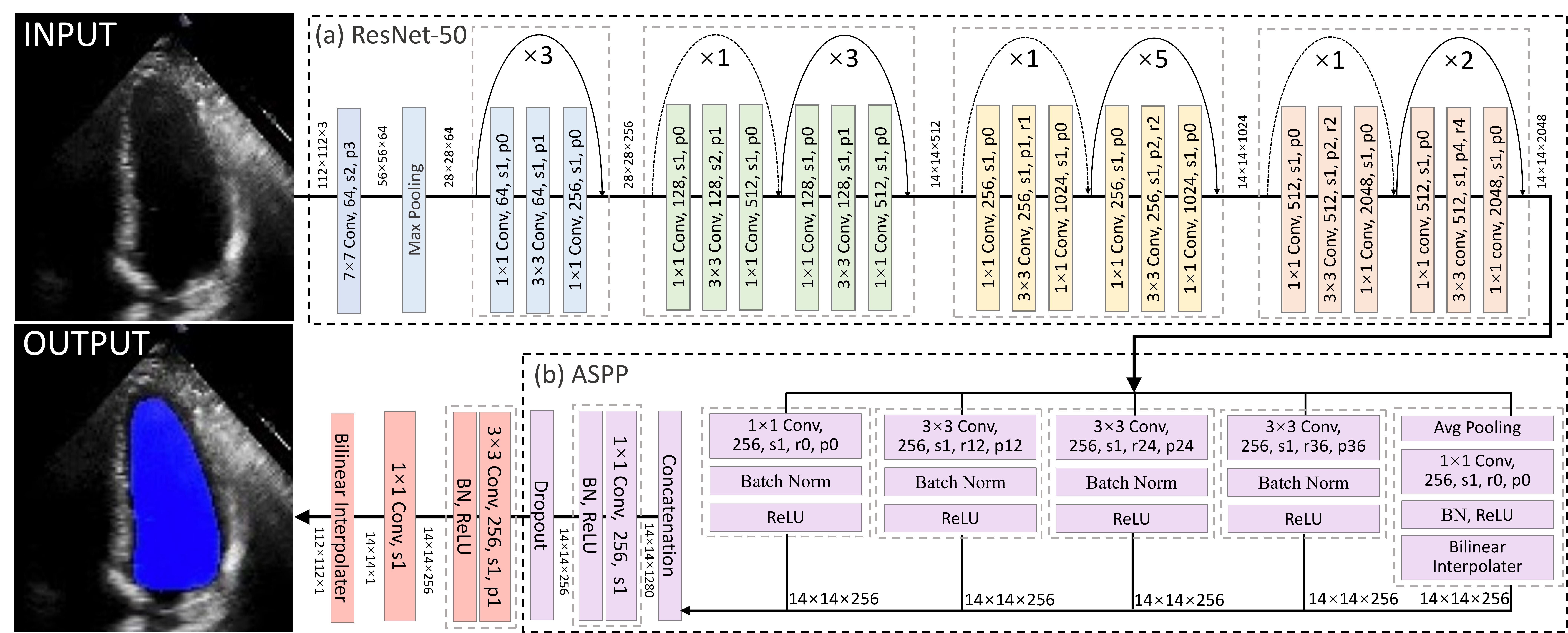}
\caption{Overall segmentation architecture. The segmentation network combined ResNet-50 (a), atrous convolutions, and atrous spatial pyramid pooling (ASPP) (b) to resample features at different scales and to capture multi-scale information. As an example, p0, r2, and s1 in the figure denotes padding = 0, atrous convolution with rate = 2, and stride = 1, respectively.}
\label{fig:3}
\end{figure}

\noindent In this project, a segmentation network, shown in Fig~\ref{fig:3}, was employed to detect the LV contours firstly, and then LV areas at ED or ES phases were computed fairly easily by counting the number of pixels from a corresponding binary mask predicted from the trained segmentation model. The proposed network combined ResNet-50, atrous convolutions, and atrous spatial pyramid pooling (ASPP) to extract feature maps and capture long range context information in the image \cite{chen2017rethinking,yu2015multi}, which was trained firstly on the training set of the Stanford dataset, and the built-in hyperparameters were tuned on its validation set. After the network had been trained, it was directly deployed to segment all frames in each video in the test set of the Stanford dataset, and then to present the trained model performance by calculating DSC between predicted masks and labelled masks at given ED and ES only. In addition, this trained model was fine-tuned in the training and validation set of the CAMUS dataset, and evaluated in its testing dataset. More details about architecture, settings, and training procedure of the model were provided in Appendix C.

\subsubsection{LV length}
\begin{figure}
\centering
\includegraphics[width=0.7\textwidth]{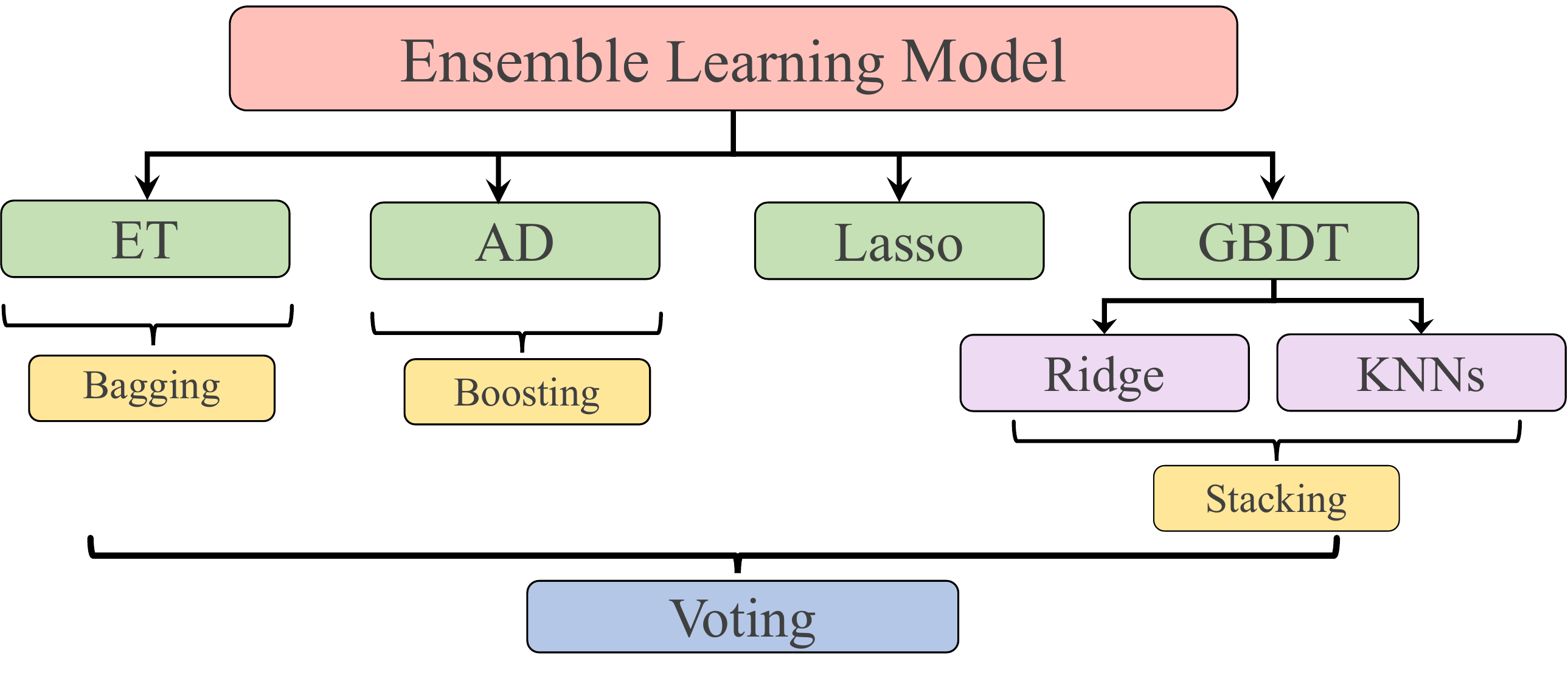}
\caption{Ensemble learning model: including Extra Tree (ET), AdaBoosting (AD), Lasso and a stacking algorithm combining Ridge, K-Nearest Neighbors (KNNs) and Gradient Boosting Decision Tree (GBDT). The predicted LV lengths from these regressors were finally ensembled by a voting mechanism.}
\label{fig:4}
\end{figure}

\noindent LV length was defined the Euclidean distance from the midpoint of the annular plane to the apex in the apical four-chamber view \cite{smistad2018fully}. Given that there is a correlation between the width, the height and the area of the polygon (representing the LV shape), as shown in Fig~\ref{fig:1}(c), a regression model based on ensemble learning (Fig~\ref{fig:4}) was developed to predict the LV length, which consists of four base regression models including Extra Trees (ET) \cite{geurts2006extremely}, Adaboosting (AD) \cite{hastie2009multi}, Lasso \cite{ranstam2018lasso} and a stacking algorithm combining Ridge \cite{pereira2016logistic}, K-Nearest Neighbors (KNNs) \cite{shakhnarovich2008nearest}, and Gradient Boosting Decision Tree (GBDT) \cite{friedman2001greedy}. This ensemble model was trained using the validation set of the Stanford dataset and their accuracy were reported on both validation and test sets of the Stanford dataset. The k-fold cross validation \cite{kohavi1995study} and the R2 score \cite{nakagawa2017coefficient} was employed to evaluate the proposed model compared with other regression models. The analysis of variance test (ANOVA) was done to prove that a significant difference between the proposed model and other comparative models \cite{long2015fully}. In addition, Pearson’s correlation coefficient (rcorr) and p-value were used to show the trained model performance on the test set of Stanford dataset \cite{dokeroglu2022comprehensive}. More details were supplied in Appendix D.

\subsubsection{ED and ES identification}
\noindent To detect all ED and ES phases in a given video, the peak detection algorithm was used taking as input the LV areas across all cardiac cycles in the video. The frame with the biggest LV area represents the ED phase, whilst the frame with the smallest LV area the ES phase. For each echocardiogram video, there are often multiple cardiac cycles. In order to identify all cardiac cycles, two parameters were defined for this algorithm. The first one was the horizontal stepsize, which was set to 20 to ensure the effective capture of all cardiac cycles (Fig~\ref{fig:5}(a)). Another parameter was the prominence value, which was set to be higher than 50\% of the global maximum minus the global minimum to assume the true peaks were located within half of the range between the maximum and minimum values (ROI 1 in Fig~\ref{fig:5}(a)). Appendix E of the supplementary file explained the parameters settings.

\subsection{Outcomes }
\noindent The main objective of this project was to determine LVEF, which is a measurement of LV systolic function utilized for heart failure phenotype classification. As a secondary outcome, this project conducted a classification task based on LVEF $<$40\% as previously calculated used all cardiac cycles to detect HFrEF samples from the test sets of both the Stanford and CAMUS datasets \cite{mcdonagh20212021, cleland2018beta}. In addition, with the computed LV areas and the identified ED as well as ES phases in section 3.3, beat-to-beat visualizer could be plotted with 1D curve where on the vertical axis it showed LV areas whilst on the horizontal axis it displayed frame numbers. This curve could be used to visualise the heart beats and carry out beat-to-beat analysis of the heart.\\

\begin{figure}
\centering
\includegraphics[width=0.98\textwidth]{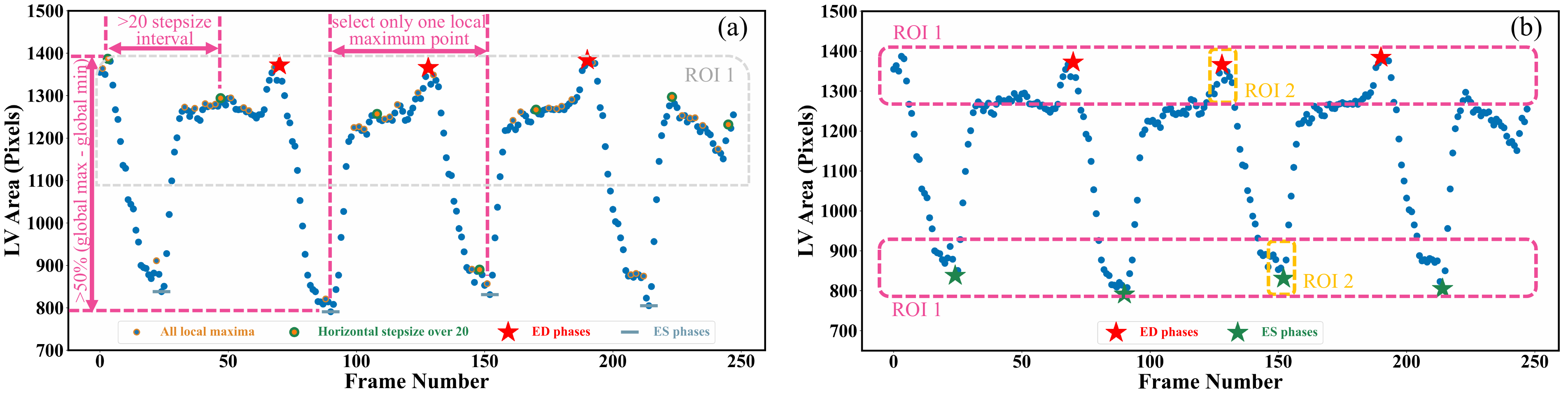}
\caption{(a) showed three scenarios used for selecting true peaks, which are identified as ED and ES phases. (b) was the improved Jeffrey’s method used to fine tune LV areas computed from segmentation. Here, three parts were averaged to compute final LV areas at ED or ES.}
\label{fig:5}
\end{figure}

\noindent \textbf{Safety and errors:} Though the proposed segmentation network was quite accurate (0.922 dice similarity coefficient on the test set), there were still errors to derive LV area due to noise. This may affect the accuracy of the LVEF, which could result in the misclassification of HF and lead to the implementation of inappropriate treatment approaches \cite{ponikowski2016task}. To further improve the performance, one method inspired by Jeffrey's method was proposed to fine tune the network prediction \cite{zhang2018fully}. Instead of directly selecting the 90\% and 10\% percentile of the left ventricular areas to serve as LV end-diastolic area and end-systolic areas, respectively, the improved Jeffrey’s method also required averaging the top 10\% ROI 1 and the top 10\% ROI 2 in Fig~\ref{fig:5}(b).

Using LV area at ED as an example, the improved Jeffrey’s method consisted of four steps: (1) Computing the LV area at ED at a specific (e.g. 2nd) cardiac cycle (indicated by the 2nd red pentagram in Fig~\ref{fig:5}(b)); (2) Computing the LV areas for each frame and sorting them according to the calculated LV areas in descending order, then selecting the top 10\% of this sorted sequence (as indicated by top ROI 1 in Fig~\ref{fig:5}(b)); (3) Sorting the frames between ED and ES within that specific (e.g., 2nd) cardiac cycle (indicated by top ROI 2 in Fig~\ref{fig:5}(b)), and then selecting the top 10\% of LV areas; (4) Averaging all these selected areas to compute the final area of LV at ED. For the LV area at ES, a similar method was used, but using descending order for sorting. The improved Jeffrey’s method was able to exclude outliers from segmentation effectively and thus improve the accuracy of predicted LVEF significantly, as shown in Fig~\ref{fig:6}(a) and (b).\\

\noindent \textbf{Analysis methods:} To evaluate the accuracy of computed LVEF, the Pearson’s correlation coefficient (rcorr) was used to show the correlation between calculated LVEF values and those provided in the respective test set \cite{dokeroglu2022comprehensive}. Additionally, the p-value was employed to measure whether the observed correlation coefficient is statistically significant. Furthermore, the student’s t-test was employed to determine whether there is a significant difference between the results from one-cycle method and those from all-cycle method. In order to evaluate the HFrEF classification, ROC curves with respective AUC values were plotted to compare the predictions with benchmark methods, which can assess the performance and discriminative ability of the classification model \cite{hanley1983method,fawcett2006introduction}. The confusion matrix was also used to visualise the performance of the proposed algorithm for showing how well the model was performing in terms of correctly predicting the target variable \cite{stehman1997selecting, powers2020evaluation}. This is particularly important because false negatives can lead to missed diagnoses or delayed treatment, highlighting its significance in medical decision-making. The confidence intervals were calculated by generating 100 bootstrapped samples and obtaining 95 percentile ranges for each prediction, aiming to estimate the level of uncertainty associated with the model's predictions.

\section{Results}
\noindent The proposed pipeline was trained and validated using the Stanford dataset (7465 and 1288 patients, respectively). The final analysis included 1270 patients, of which 8\% (106) had LVEF $<$ 40\%. Iteration and external validation used the CAMUS dataset of 200 patients, of which 66 (33\%) had LVEF $<$ 40\%, 62 (31\%) were women and average age was 64.9 years. Image quality for echocardiography in the CAMUS dataset was reported as good in 113 patients (57\%), adequate in 65 patients (32\%), and poor in 22 patients (11\%). Further details on patient characteristics are summarized in Table S1 in Appendix B.
\subsection{Accuracy of Automated LVEF Calculation }
\noindent The automated method to compute LVEF given in formulation~\ref{eq:2} was assessed in three experiments, based on the segmentation network and LV length model that were trained and elaborated upon in the Appendix C and D of the supplementary file. \\

\begin{figure}
\centering
\includegraphics[width=0.98\textwidth]{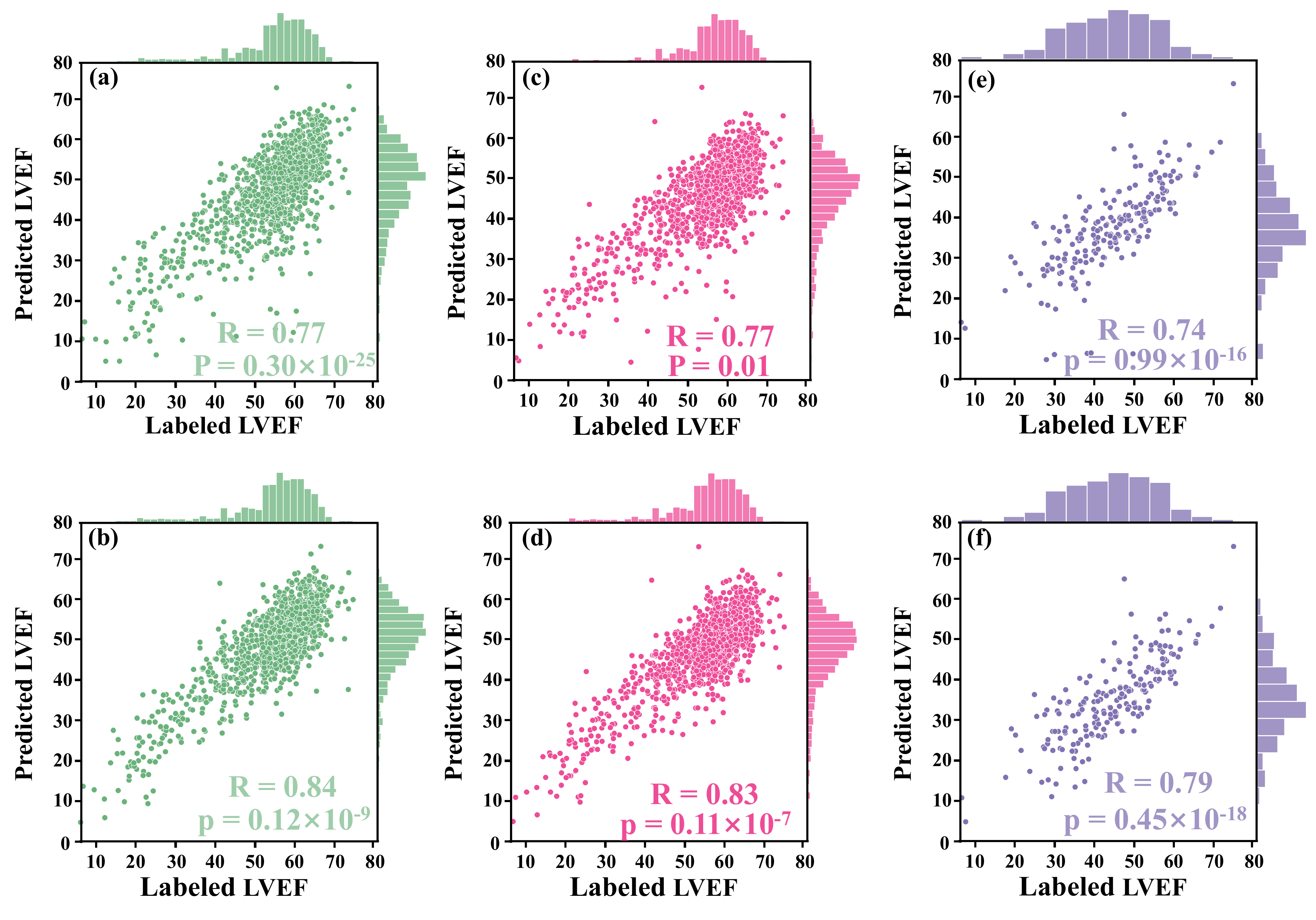}
\caption{Correlation plots. (a)-(d) were results from the Stanford dataset, while (e)-(f) from the CAMUS dataset. (a) showed the correlation between LVEF values derived from segmentation results directly and those labelled by an experienced clinician. (b) showed the correlation between LVEF values derived from the proposed Jeffrey’s method and those labelled by the clinician. (c) showed the correlation between LVEF values computed from a single cardiac cycle and labelled LVEF values. (d) showed the correlation between LVEF values computed from all cardiac cycles and labelled LVEF values. (e) showed the correlation between LVEF values derived from fine-tuned segmentation results and labelled LVEF values. (f) showed the correlation between LVEF values derived from the improved Jeffrey’s method and labelled LVEF values.}
\label{fig:6}
\end{figure}

\noindent\textbf{Experiment 1:} The alternative hypothesis was that the Jeffrey’s method proposed in Section 3.4 could improve the performance of computing LVEF. For this, the ED and ES frames provided in the test set of the Stanford dataset were used. For each sample in the test set, LV Lengths were predicted by the proposed voting ensemble learning model already trained in Section 4.2. LV areas were predicted by two methods: one was to deploy the trained network to segment their ED and ES frames and then count the number of pixels in the segmentation masks, and the other one was the improved Jeffrey’s method. As shown in Fig~\ref{fig:6}(a) and (b), these two sub-figures showed that the LVEF values derived from segmentation directly had a rcorr value of 0.77 (p-value $<$ 0.0001, 95\% CI 0.74 to 0.80) with respect to these LVEF values provided in the test set. The correlation could be boosted to 0.84 (p-value $<$ 0.0001, 95\% CI 0.82 to 0.86) when using the improved Jeffrey’s method to compute LV areas. This experiment showed that it was necessary to fine tune LV areas after segmentation using the proposed Jeffrey’s method, which improves the accuracy of resulting LVEF with t-value less than 0.0001.\\

\noindent \textbf{Experiment 2:} The alternative hypothesis was that LVEF computed by averaging across all cardiac cycles (i.e., our equation (2) where N $>$ 1) was more accurate than that from only single cardiac cycle (i.e., the equation (2) where N $=$ 1), where the reference was human estimates of LVEF. First, the proposed peak detection algorithm was used to identify all ED and ES phases in a given echocardiogram video from the test set of the Stanford dataset. For the former method, the first paired ED and ES frames were selected as a cycle and then to compute LVEF. For the latter method, all identified cycles were used to compute an averaged LVEF value using (2) for this video (85\% of the videos contain more than three cardiac cycles). As shown in Fig~\ref{fig:6}(c) and (d), the LVEF values derived from single cycle (rcorr $=$ 0.77, p-value $=$ 0.01, 95\% CI 0.75 to 0.80) were less accurate than those derived from all cycles (rcorr $=$ 0.83, p-value $<$ 0.0001, 95\% CI 0.81 to 0.85), when referring to these LVEF values provided in the test set (t-value $<$ 0.0001). Furthermore, if the second cycle was selected to compute LVEF, their respective rcorr value could be boosted to 0.78 (p-value $<$ 0.0001, 95\% CI 0.78 to 0.80), still inferior to the proposed all-cycle method (t-value $<$ 0.0001). \\

\noindent\textbf{Experiment 3:} The alternative hypothesis was that performance of the model would be retained in an external dataset (test set of the CAMUS dataset). To predict LV areas, the segmentation network trained from the Stanford dataset had been fine-tuned on the training set of CAMUS dataset and then it was deployed on the test set of CAMUS. To predict LV lengths, the voting ensemble learning model trained from the Stanford dataset was deployed directly on the test set of CAMUS. As shown in Fig~\ref{fig:6}(e) and (f), it could be seen that the rcorr value was improved from 0.74 (p-value $<$0.0001, 95\% CI 0.68 to 0.78) to 0.79 (p-value $<$0.0001, 95\% CI 0.74 to 0.84) before and after applying for the proposed Jeffrey’s method. 

\subsection{Classification of patients with HFrEF }
\begin{figure}
\centering
\includegraphics[width=0.98\textwidth]{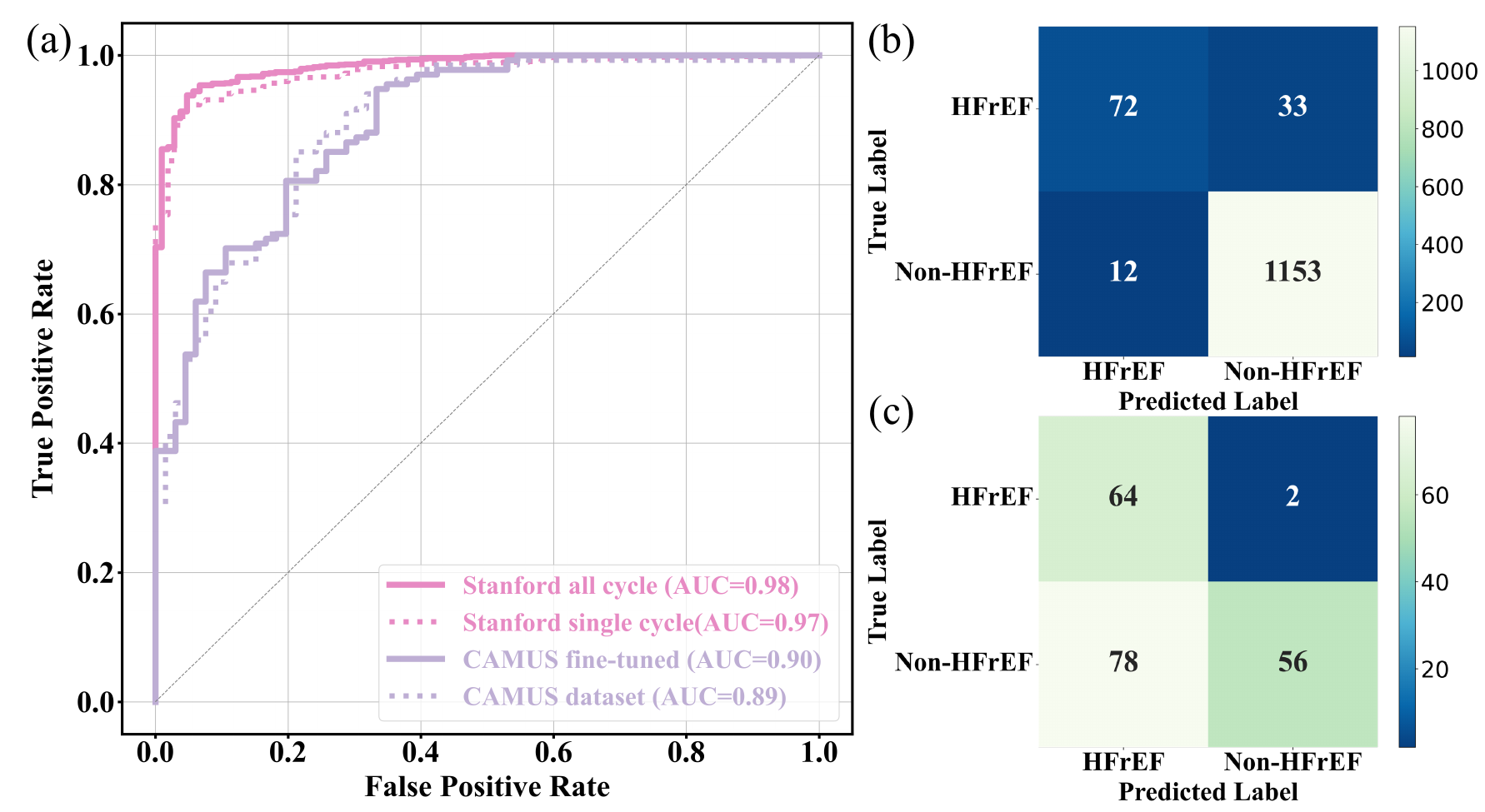}
\caption{HFrEF assessment results. (a) shows the ROC curves of different methods, each having an AUC value. (b) and (c) are the confusion matrices computed from the Stanford and CAMUS datasets, respectively.}
\label{fig:7}
\end{figure}

\noindent Current HFrEF terminology was used as guidance to detect HFrEF samples from the test sets of both Stanford and CAMUS datasets based on their LVEF predicted in Section 4.3.1. ROC curves were plotted, and their AUC values were computed in Fig~\ref{fig:7}(a). Among these curves (see the plot legend), the first two were obtained on the Stanford dataset whilst the last two on the CAMUS. The proposed all-cycle method achieved AUC value with 0.98 (95\% confidence interval 0.97 to 0.99) in the internal validation (Stanford dataset). On external validation using the CAMUS dataset, AUC was 0.90 (95\% confidence interval 0.88 to 0.91), as shown in Table~\ref{table:1}.

In addition, the confusion metric was presented to further evaluate the accuracy of the proposed methods. Fig~\ref{fig:7}(b) and (c) showed the results from the test set of Stanford dataset and CAMUS, respectively. For the Stanford dataset, there were 1270 samples in its test set, where 97\% of samples were classified correctly. There were 12 that were not HFrEF samples, but the classifier classified them as HFrEF. There were 33 HFrEF samples, but the classifier classified them as non-HFrEF. With regards to the confusion metric for CAMUS, the proposed method predicted 78 non-HFrEF as HFrEF patients, but only two with HFrEF were mistaken as non-HFrEF.

\begin{table}[htbp]
    \centering
    \caption{HFrEF assessment results using AUC values with a confidence interval of 95\%.}
    \begin{tabular}{lcc}
        \hline
        & Stanford & CAMUS \\
        \hline
        Single Cycle & 0.97 (0.96-0.98) & 0.89 (0.87-0.91) \\
        \cline{2-3}
        Average Cycle & 0.98 (0.97-0.99) & 0.90 (0.88-0.91) \\
        \hline
    \end{tabular}
    \label{table:1}
\end{table}

\subsection{Beat-to-Beat Visualizer}
\begin{figure}
\centering
\includegraphics[width=0.98\textwidth]{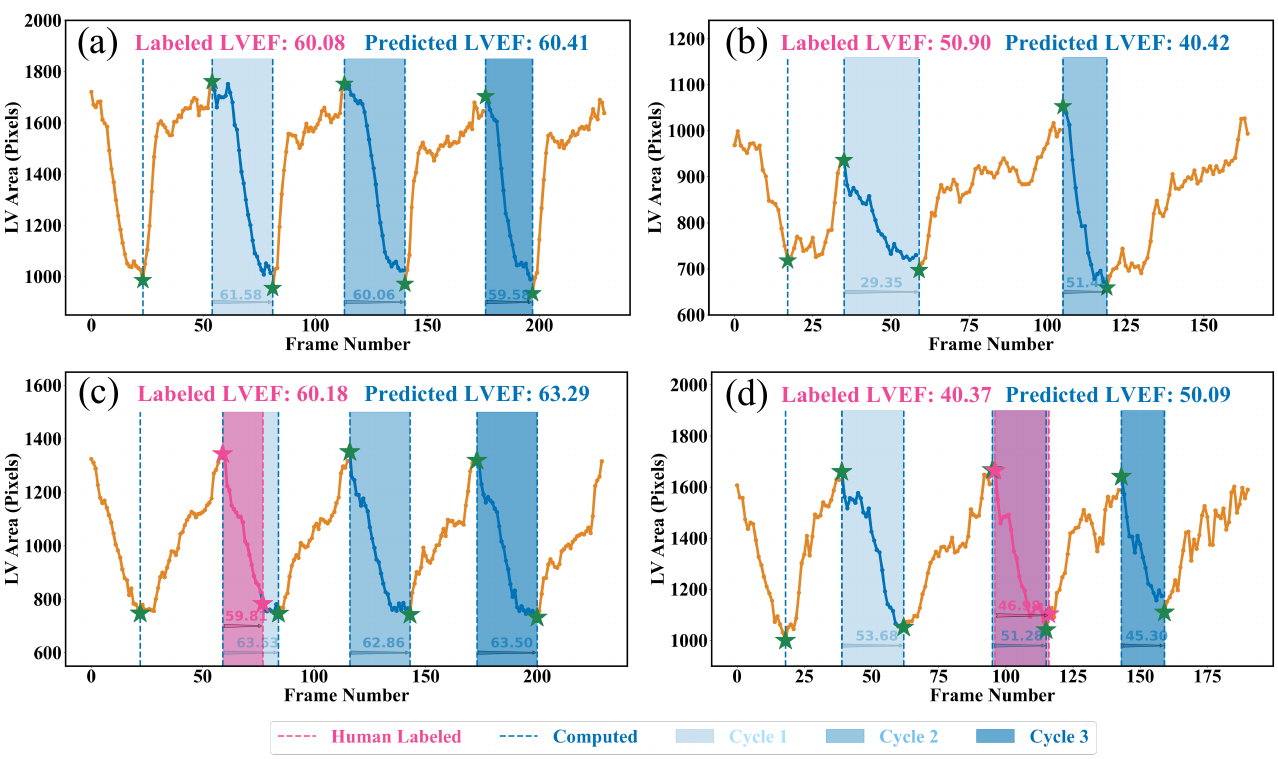}
\caption{Beat-to-beat analysis. (a) and (c) shown two samples with normal sinus rhythm. (b) in a patient with atrial fibrillation. (c) and (d) showed human labelled ED and ES were not exactly at peak or bottom positions.}
\label{fig:8}
\end{figure}

\noindent Beat-to-Beat visualizer provided as the outputs for diagnostic purposes, in addition to quantitative results given in the previous sections. Based on the computed LV areas and the identified ED as well as ES phases, two Beat-to-Beat visualizers were presented in Fig~\ref{fig:8}(a) and Fig~\ref{fig:8}(b), which were used to overview of LV volumes across all cardiac cycles and provide an instantaneous summary of beat-to-beat volumetric differences as a result of sinus or pathological arrhythmias. In Fig~\ref{fig:8}(a), there was a similar gap between ED and ES frames, which was the sample with a normal sinus rhythm in heart beats. Fig~\ref{fig:8}(b) was a sample marked as a patient with atrial fibrillation by the dataset publisher. This figure showed that the sample had irregular heartbeats, and the gap between ED and ES frames varied across all cardiac cycles. These examples provided a visualisation for hearts having different conditions.

\section{Discussion}
\noindent This project proposed a novel pipeline method to assess cardiac function that achieved state-of-the-art results. It involved training a weakly supervised algorithm to identify the left ventricle using expert tracings, followed by using an ellipsoid single-plane model to determine LVEF values. This pipeline outperformed previous attempts that relied on segmentation-based deep learning methods \cite{zhang2018fully}. Furthermore, its performance in predicting the LVEF values was robust when applied to an external dataset of echocardiogram sequences from an independent medical centre. As a result, this pipeline could have the potential to assist clinicians in achieving a more precise and reproducible assessment of cardiac function, and can have the capability to identify subtle changes in LVEF beyond the precision of human readers. \\

\noindent One difference between the proposed pipeline and human evaluation was the method of calculating LVEF, where the pipeline was based on beat-to-beat evaluation across numerous cardiac cycles, while with the typical clinical approach is to take just one representative beat. The process of tracing three or five beats is not commonly performed in routine practice due to the labour-intensive and time-consuming nature of the task. By automating the segmentation task, the proposed pipeline reduced the labour involved in assessing cardiac function and allowed for more frequent and accurate evaluations.

Two examples from the test set of the Stanford dataset were presented in Fig~\ref{fig:8}(c) and Fig~\ref{fig:8}(d) to further explain the reason of using all-cycle method. As can be seen, there were three cardiac cycles in Fig~\ref{fig:8}(c), with three LVEF values being 60.53\%, 62.86\%, and 63.50\%, respectively. In this case, calculating LVEF from any cycle would not make a significant difference. In Fig~\ref{fig:8}(d), there were also three cycles, with the corresponding LVEF values being 53.68\%, 51.28\%, and 45.30\%, respectively. If using the third cycle to compute LVEF, it would end up with identifying this sample with HFmrEF, which resulted in a true negative classification. Using the all-cycle method, the LVEF value was 50.09\%, with which it was able to classify this sample correctly as HFpEF. Therefore, some recent studies based on only single-cycle information rather than all-cycle information might lead to reduced reliability and accuracy in diagnosing patients with systolic heart failure \cite{leclerc2019deep,zhang2018fully,dong2016left,smistad2020real,thavendiranathan2012feasibility}.\\

\noindent Another difference was that the pipeline relied on the machine to identify LV contours and ED as well ES frames, which had the capability of computing LVEF more accurately. For example, in Fig~\ref{fig:8}(d), with pink ED and ES, the LVEF value is 46.98\% (HFmrEF), whilst with the corresponding green ED and ES, the LVEF value was 51.28\% (HFpEF). According to the Stanford dataset publisher, this sample should have a LVEF value above 50\% \cite{ouyang2020video}. Clearly, this method computed a correct LVEF, proving the effectiveness of the proposed peak detection algorithm, while labelling ED and ES incorrectly would result in incorrect LVEF. This means the ground truth LVEF values used to train the network may be already inaccurate for some regression method, due to the fact that the selection of ED and ES frames might be incorrect and that only one cycle was used to calculate LVEF in practice rather than using three or five consecutive cardiac cycles as per the ASE recommendation. Therefore, for some regression methods used these incorrect labels to train models, their prediction and evaluation accuracy could be degraded and biased \cite{ouyang2020video, ghorbani2020deep, jiang2019deep,kusunose2020deep}. However, the automated methods in this study had no such issues and therefore were better than direct regression methods.\\

\noindent One limitation of the validation was the relatively small sample size from CAMUS dataset (only 200 samples used for fine tuning the network). However, the results of the LVEF were still robustly accurate when applying this learned model to CAMUS dataset originating from a different site and time interval. Another limitation was the inability to use the Simpson’s biplane method (measurement of LVEF using both A4C and apical 2-chamber views), as recommended by ASE and EACVI, due to the Stanford Echo-Dynamic dataset only providing A4C views \cite{stamm1982two, fonarow2016left}. Instead, the area-length formulation was employed based on the ellipsoid single-plane model, which still showed an excellent correlation with human labelled LVEF calculated with Simpson’s biplane (r$=$0.99; p$<$ 0.0001; mean absolute error 4.4\%). Furthermore, the proposed approach could easily be modified to take into account the biplane method of LVEF calculation, with LV areas for both views derived from two separate segmentation ACNN and the improved Jeffrey’s method, while LV length could be derived from the novel ensemble learning model. 

\section{Conclusion}
\noindent In this project, a new pipeline method was proposed to assess cardiac function based on only Apical 4 chamber cines, which could not only provide quantitative results, such as left ventricular ejection fraction, but also present left ventricular contours and beat-to-beat visualizers for cardiologists to visually view the samples while making diagnoses. Additionally, the study highlighted the importance of following the ASE and EACVI recommendation of averaging three or five cycles to obtain a more precise assessment.

\section{Data Availability Statement}
\noindent The original datasets presented in the study are found at \url{https://stanfordaimi.azurewebsites.net/datasets} and \url{https://www.creatis.insa-lyon.fr/Challenge/camus/databases.html}, further inquiries can be directed to the corresponding author. 

\section{Conflict of Interest}
\noindent All authors have completed the ICMJE uniform disclosure form and declare: DK reports grants from the National Institute for Health Research (NIHR CDF-2015-08-074 RATE-AF; NIHR130280 DaRe2THINK; NIHR132974 D2T-NeuroVascular; NIHR203326 Biomedical Research Centre), the British Heart Foundation (PG/17/55/33087, AA/18/2/34218 and FS/CDRF/21/21032), the EU/EFPIA Innovative Medicines Initiative (BigData@Heart 116074), EU Horizon (HYPERMARKER 101095480), UK National Health Service -Data for R\&D-Subnational Secure Data Environment programme, UK Dept. for Business, Energy \& Industrial Strategy Regulators Pioneer Fund, the Cook \& Wolstenholme Charitable Trust, and the European Society of Cardiology supported by educational grants from Boehringer Ingelheim/BMS-Pfizer Alliance/Bayer/Daiichi Sankyo/Boston Scientific, the NIHR/University of Oxford Biomedical Research Centre and British Heart Foundation/University of Birmingham Accelerator Award (STEEER-AF). In addition, he has received research grants and advisory board fees from Bayer, Amomed and Protherics Medicines Development, all outside the submitted work. JD reports grants from the British Heart Foundation University of Birmingham Accelerator (AA/18/2/34218) and the Korea Cardiovascular Bioresearch Foundation (CHORUS Seoul 2022)

\section{Author Contributions}
\noindent Conceptualization, methodology, experiments, and writing original draft preparation: YZ and JD. Review and editing: JD, BL, KB, DB, AT, AK, GG and DK. Funding acquisition: JD and DK. All authors have read and agreed to the published version of the manuscript. 

\section{Acknowledgement}
\noindent The card\textit{AI}c team at the University of Birmingham and University Hospitals Birmingham NHS Foundation Trust have received support and funding from the NIHR Birmingham Biomedical Research Centre (NIHR203326), MRC Health Data Research UK (HDRUK/CFC/01), NHS Data for R\&D Subnational Secure Data Environment programme (West Midlands), the British Heart Foundation University of Birmingham Accelerator (AA/18/2/34218) and the Korea Cardiovascular Bioresearch Foundation (CHORUS Seoul 2022). The opinions expressed in this paper are those of the authors and do not represent any of the listed organizations; None of the organizations had any role in design or conduct of the study (including collection, analysis and interpretation of the data) or any involvement in preparation, review or approval of the manuscript.

\bibliographystyle{elsarticle-num}
\bibliography{main}







\end{document}